# A Fuzzy Logic Based Certain Trust Model for E-Commerce

Kawser Wazed Nafi, Tonny Shekha Kar, Md. Amjad Hossain, M. M. A. Hashem
Computer Science and Engineering Department,
Khulna University of Engineering and Technology
Khulna, Bangladesh
kwnafi@yahoo.com, tulip0707051@gmail.com, amjad_kuet@yahoo.com, mma.hashem@outlook.com

*Abstract*— **Trustworthiness especially for service oriented system is very important topic now a day in IT field of the whole world. There are many successful E-commerce organizations presently run in the whole world, but E-commerce has not reached its full potential. The main reason behind this is lack of Trust of people in e-commerce. Again, proper models are still absent for calculating trust of different e-commerce organizations. Most of the present trust models are subjective and have failed to account vagueness and ambiguity of different domain. In this paper we have proposed a new fuzzy logic based Certain Trust model which considers these ambiguity and vagueness of different domain. Fuzzy Based Certain Trust Model depends on some certain values given by experts and developers. can be applied in a system like cloud computing, internet, website, e-commerce, etc. to ensure trustworthiness of these platforms. In this paper we show, although fuzzy works with uncertainties, proposed model works with some certain values. Some experimental results and validation of the model with linguistics terms are shown at the last part of the paper.**

*Keywords-Certain Trust; Fuzzy Logic; Probabilistic Logic; Subjective Logic;E-Commerce Trust Model*

## I. INTRODUCTION

TRUST is a well-known concept in everyday life and often serves as a basis for making decisions in complex situations. There are numerous approaches for modeling trust concept in different research fields of computer science, e.g., virtual organizations, mobile and P2P networks, and E-Commerce. The sociologist Diego Gambetta has provided a definition, which is currently shared or at least adopted by many researchers. According to him "Trust is a particular level of the subjective probability with which an agent assesses that another agent or group of agents will perform a particular action, both before he can monitor such action and in a context in which it affects its own action" [1].

E-commerce is seen as an extension of mail and phone order transactions and is gaining popularity. In E-commerce, business transactions are no longer bound to physical existence, geographic boundaries, time differences or distance barriers. Han and Noh [2] found that several critical failure factors of E-commerce need to be addressed seriously by the industry to ensure that E-commerce usage will continue to grow. The findings are mainly on the dissatisfaction of customers on the unstable E-commerce systems, a low level of personal data security, inconvenience systems, disappointing purchases, unwillingness to provide personal details and mistrust of the technology. Indeed, customers may doubt the quality of the goods as they may find it difficult to engage in a transaction without proper testing, seeing and touching the products.

To succeed in the fiercely competitive e-commerce marketplace, businesses must become fully aware of Internet security threats, take advantage of the technology that overcomes them, and win customers' trust. Eighty-five percent of Web users surveyed reported that a lack of security made them uncomfortable sending credit card numbers over the Internet. The merchants who can win the confidence of these customers will gain their loyalty—and an enormous opportunity for expanding market share [3].

In person-to-person transactions, security is based on physical cues. Consumers accept the risks of using credit cards in places like department stores because they can see and touch the merchandise and make judgments about the store. On the Internet, without those physical cues, it is much more difficult to assess the safety of a business. Also, serious security threats have emerged. By becoming aware of the risks of Internet-based transactions, businesses can acquire technology solutions that overcome those risks: Spoofing, Unauthorized Disclosure, Unauthorized Action, Eavesdropping and Data Alteration.

In this paper, we are going to propose a newer trust model which is based on fuzzy logic and probabilistic logic. Many trust model developers used "Subjective Logic" for their trust mode. But this subjective model has some problems like: fail to work with uncertain values, lacks of clear mathematical results, etc. Proposed model will solve the problems with "Subjective Logic". In this paper, the mathematical equations are designed such a way that clients and customers can easily understand the output of the model and can take their decision easily, because humans can easily understand fuzzy model's output.

The whole paper is divided in following sections: section II describes the related E-Commerce work of the proposed model, Section III describes the proposed model and its output for different modules, section IV discusses the experimental results which are worked out in Lab and advantages of the proposed model and section V discusses about the conclusion part of the whole proposed model.

## II. RELATED WORK

Several number of ways are there for modeling (un-)certainty of trust values in the field of trust modeling in Cloud computing and internet based marketing sys-tem.[4] But, these models have less capability to derive trustworthiness of a system which are based on knowledge about its components and subsystems. Fuzzy logic was used to provide trust in Cloud computing. Different types of attacks and trust models in service oriented systems, distributed system and so on are designed based on fuzzy logic system [5]. But it models different type of uncertainty known as linguistical uncertainty or fuzziness. In paper [6], a very good model for E-commerce, which is based on fuzzy logic, is presented. But, this model also works with uncertain behavior. Belief theory such as Dempster-Shafer theory was used to provide trust in Cloud computing [7]. But the main drawback of this model is that the parameters for belief, disbelief and uncertainty are dependent on each other. It is possible to model uncertainty using Bayesian probabilities[8] which lead to probability density functions e.g., Beta probability density function. It is also possible to apply the propositional standard operators to probability density functions. But this leads to complex mathematical operations and multi-dimensional distributions which are also hard to interpret and to visualize. An enhanced model recently being developed for using in Cloud computing is known as Certain Trust. This model evaluates propositional logic terms that are subject to uncertainty using the propositional logic operators AND, OR and NOT[1].

Manchala [9] proposes a model for the measurement of trust variables and the fuzzy verification of E-Commerce transactions. He highlights the fact that trust can be determined by evaluating the factors that influence it, namely risk. He defines cost of transaction, transaction history, customer loyalty, indemnity and spending patterns as the trust variables. But he fails to solve the following problems of E-Commerce: - Suitable variables for outputs, establish relations between variables and fails to support theoretical logics for E-Commerce trust models. Jøsang [10] also works with trust models and work with "Subjective Logic", but this model is fully depended with uncertainty. S.Nefti proposed a model which solves the problem of Manchala and Jøsang, but fails to solves problems related with uncertainty and can't show behavioral probability of a E-Commerce based company.

## III. PROPOSED TRUST MODEL FOR E-COMMERCE

Certain Trust Model was constructed for modeling those probabilities, which are subject to uncertainty. This model was designed with a goal of evidence based trust model. Moreover, it has a graphical, intuitively interpretable interface [1] which helps the users to understand the model (Figure 1). The representational model focuses on two crucial issues

a) How trust can be derived from evidence considering context-dependent parameters?

b) How trust can be represented to software agents and human users?

For the first one, a relationship between trust and evidence is needed. For this, they had chosen a Bayesian approach. It is because it provides means for deriving a subjective probability from collected evidence and prior information [1]. At developing a representation of trust, it is necessary to consider to whom trust is represented. It is easy for a software component or a software agent to handle mathematical representations of trust. For it, Bayesian representation of trust is appropriate. The computational model of Certain Trust proposes a new approach for aggregating direct evidence and recommendations. In general, recommendations are collected to increase the amount of information available about the candidates in order to improve the estimate of their trustworthiness. This recommendation system needs to be integrated carefully for the candidates and for the users and owner of cloud servers. This is called robust integration of recommendations. In order to improve the estimate of the trustworthiness of the candidates, it is needed to develop recommendation system carefully. This is called robust integration of recommendations [1].

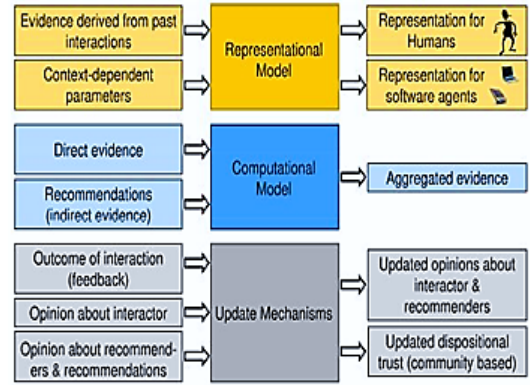

Figure 1. Block diagram of Trust models

Three parameters used in certain logic: average rating t, certainty c, initial expectation f. The average rating t indicates the degree to which past observations support the truth of the proposition. The certainty c indicates the degree to which the average rating is assumed to be representative for the future. The initial expectation f expresses the assumption about the truth of a proposition in absence of evidence [1].

The equations for these parameters are given below:-

Equation for average rating, $t = \begin{cases} 0.5 & if\ r+s = 0 \\ \frac{r}{r+s} & else \end{cases}$ (1)

Here, r represents number of positive evidence and s represents number of negative evidence defined by the users or third person review system.

Equation for certainty, $c = \frac{N.(r+s)}{2.w.(N-(r+s))+N.(r+s)}$ (2)

Here, w represents dispositional trust which influences how quickly the final trust value of an entity shifts from base trust value to the relative frequency of positive outcomes and N represents the maximum number of evidence for modeling trust. Using these parameters the expectation value of an opinion $E(t,c,f)$ can be defined as follows:

$$E(t,c,f) = t*c + (1-c)*f \qquad (3)$$

The parameters for an opinion o = (t, c, f) can be assessed in the following two ways: direct access and Indirect access. Certain Trust evaluates the logical operators of propositional logic that is AND, OR and NOT. In this model these operators are defined in a way that they are compliant with the evaluation of propositional logic terms in the standard probabilistic approach. However, when combining opinions, those operators will especially take care of the (un)certainty that is assigned to its input parameters and reflect this (un)certainty in the result. The definitions of the operators as defined in the CTM are given in the table 1.

TABLE I. DEFINITION OF OPERATORS

| | |
|---|---|
| OR | $c_{A \vee B} = c_A + c_B - c_A c_B - \frac{cA(1-cB)fB(1-tA)+(1-cA)cBfA(1-tB)}{fA+fB-fAfB}$ <br> $t_{A \vee B} = \begin{cases} \frac{1}{c_{A \vee B}}(c_A t_A + c_B t_B - c_A c_B t_A t_B) & \text{if } c_{A \vee B} \neq 0 \\ 0.5 & \text{else} \end{cases}$ <br> $f_{A \vee B} = f_A + f_B - f_A f_B$ |
| AND | $c_{A \wedge B} = c_A + c_B - c_A c_B - \frac{(1-cA)cB(1-fA)tB+cA(1-cB)(1-fB)tA}{1-fAfB}$ <br> $t_{A \wedge B} = \begin{cases} \frac{1}{c_{A \wedge B}}(c_A c_B t_A t_B + \frac{cA(1-cB)(1-fA)fBtA+(1-cA)cBfA(1-fB)tB}{1-fAfB}) & \text{if } c_{A \wedge B} \neq 0 \\ 0.5 & \text{else} \end{cases}$ <br> $f_{A \wedge B} = f_A f_B$ |
| NOT | $t'_A = 1 - t_A$, $c'_A = 1 - c_A$ and $f'_A = 1 - f_A$ |

With the help of these parameters and operators derived from Certain Trust, we have defined two new parameters, Trust T and behavioral probability, P [11]. Trust T is calculated from certainty c and average rating t. the equation is:

$$\text{Trust, } T = \frac{c*t}{High\, scaling\, value\, of\, rating} * 100\% \quad (4)$$

Here, High scaling value of rating means the upper value of the range of rating.

Calculating T, we have applied FAM rule of fuzzy logic for creating a relation between certainty c and average rating t. Trust T represents this relation in percentage such a way that the quality of the product can easily be understood. Another parameter, behavioral probability, P, represents how the present behavior of the system varies from its initially expected value and it is proposed with the help of probabilistic logic. It may be less, equal or higher than the initial expectation given by the system developer or the manager of the office. The equation for P is:

$$\text{Behavioral probability, } P = \frac{(T)-f}{f} * 100\% \quad (5)$$

If T<f, lower probability to show expected behavior
If T>f, higher probability to show expected behavior
If T=f, balanced with the expected behavior

According to Gaussian, the membership function for fuzzy input sets depends on two types of parameter, standard deviation **σ** and mean **c.** The equation for membership function is:-

$$f(x; \sigma; c) = \exp\left(\frac{-(x-c)^2}{2\sigma^2}\right) \quad (6)$$

Fuzzy inference is the process of formulating the mapping from a given input to an output using fuzzy logic. The mapping then provides a basis from which decisions can be made, or patterns discerned. There are two concepts of fuzzy logic systems. They are: - linguistic variables and fuzzy if then else rule. The linguistic variables' values are words and sentences where if then else rule has two parts; antecedents and consequent parts which contain propositions of linguistic variables. Numerical values of inputs $x_i \epsilon U_i (i = 1,2 \ldots n)$ are fuzzified into linguistic values, $F1, F2 \ldots Fn$. Here $F_i$ denotes the universe of discourse $U = U1 * U2 * \ldots \ldots * Un$. The output linguistic variables are $G1, G2, \ldots \ldots, Gn$. The if-then-else rule can be defined as: -

$$R^{(j)}: IF\ x_i\ F_1^j\ and \ldots \ldots and\ x_n\ F_n^j\ THEN\ y\ G^j. \quad (7)$$

The trust relationships among customers and vendors are hard to assess due to the uncertainties involved. Two advantages of using fuzzy-logic to quantify trust in E-commerce applications are: (1) Fuzzy inference is capable of quantifying imprecise data and quantifying uncertainty in measuring the trust index of the vendors. (2). Fuzzy inference can deal with variable dependencies in the system by decoupling dependable variables. The general trust model proposed by s.nemfi [12] is composed of five modules. Four modules will be used to quantify the trust measure of the four factors identified in our trust model (Existence, Affiliation, Policy and Fulfilment). The fifth module will be the final decision maker. Figure II describes general trust model for E-commerce.

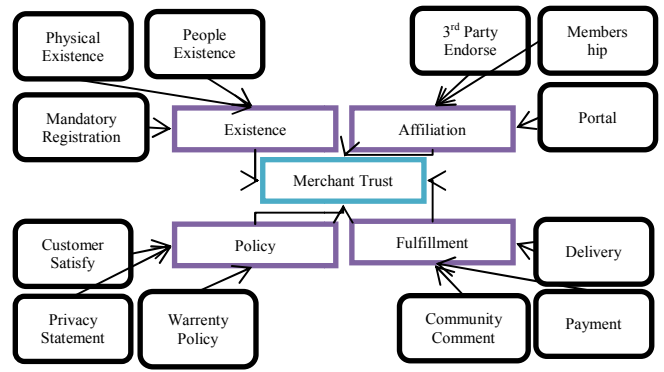

Figure 2. Trust Model for E-commerce

The inputs of the first module are Physical Existence, People Existence and Mandatory registration and the output of the affiliation module. For module 2, the inputs are Third Party Endorsement, Membership and Portal and output is affiliation variables. Module 3 inputs are Customer Satisfaction, Privacy and Warranty for Policy. Finally, the fourth module inputs are represented by Delivery, Payment Methods and Community Comment. The Finale module combines the four modules which are Existence, Fulfillment, Policy and the Affiliation for producing the final output Merchant Trust. Trust values of every element of each module are calculated with the help of parameter c and t and then average values of them work as output of each model. We divide the classes of all input and output variables into five parts according to acceptable degrees of survey.

Fuzzy specification rules, according to equation 8, used for Existence module are given below:-

1. If (People_Edistence is Very_Low) and (Physical_Existence is Very_Low) and (Mendatory_Registration is Very_Low) then (Existence is Very_Low) (1)
2. If (People_Edistence is Low) and (Physical_Existence is Very_Low) and (Mendatory_Registration is Very_Low) then (Existence is Very_Low) (1)
..................................................................
130. If (People_Edistence is Very_High) and (Physical_Existence is Very_High) and (Mendatory_Registration is Very_High) then (Existence is Very_High) (1)

With the help of fuzzy linguistic variables, we get the final output of Existence modules given in Figure 3.

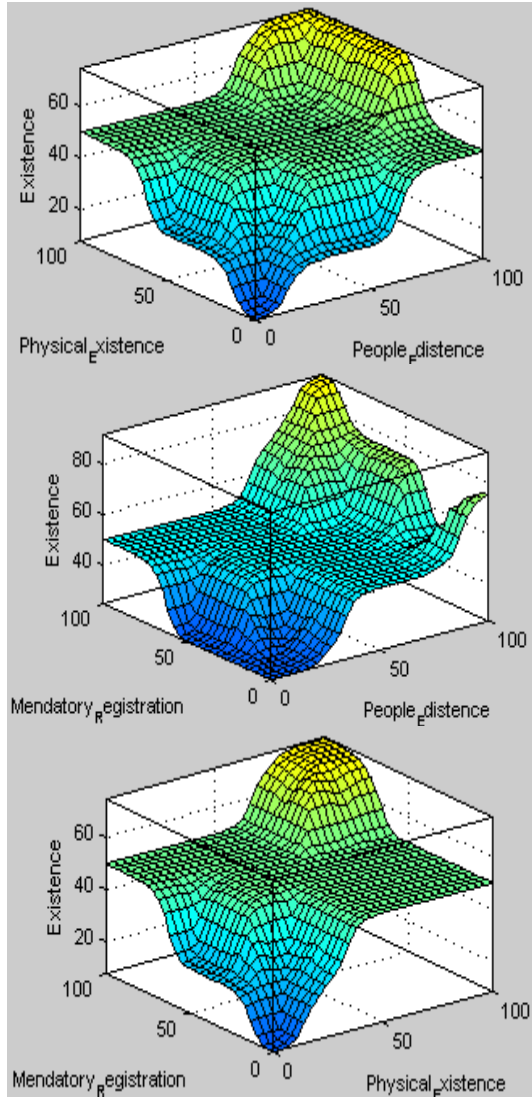

Figure 3. Mapping Surface for Existence Module

After getting fuzzy outputs from 4 modules of trust model, we have used them as input for our final module, Merchant trust and have got the membership functions shown in Figure 5 and outputs for different variables shown in Figure 4.

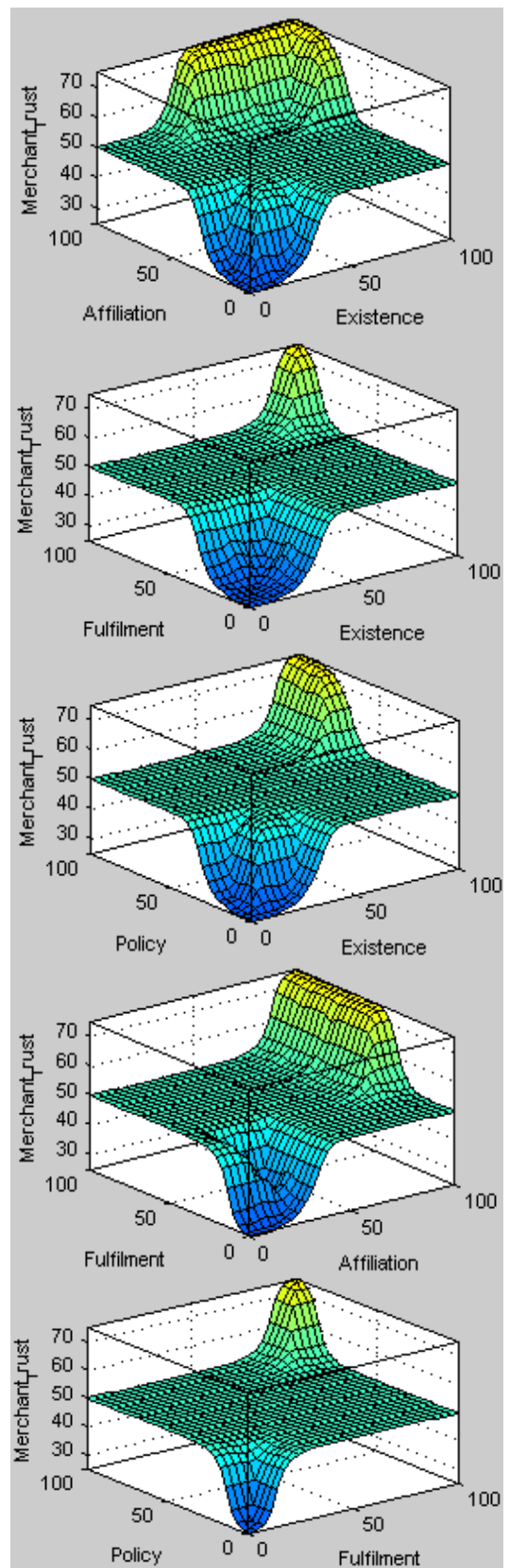

Figure 4. Mapping Surface for Merchant Trust Module

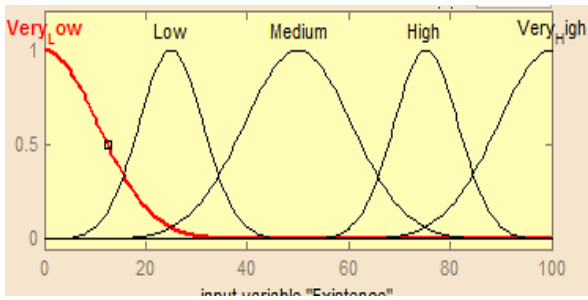

Figure 5. Membership functions for Final Output Merchant Trust

IV. EXPERIMENTAL RESULTS

In this section we are going to show the effects of Fuzzy Based Certain Trust model in the field of E-Commerce Architecture. We have run this model in our lab for several times and have got different types of trust values for different environments. We are going to show two scenarios consisting 100 people each and trying to show the evaluating process of our proposed model.

*A. Case Study 1:*

Let us consider a company, name A, running an online business following our E-Commerce structure. According to Certain Trust Model's operators defined in equation 3, 4, and 5, we know that, the input for this model is r, s, f and w. Let, for any time the input values are r=5, s=2, f=0.5 and w=1 where, number of evidences are N=7.Then, the output values are:- average rating t = 0.714 and c=0.724. and then, E = 0.65. Now, for mapping it to our proposed model, we need to modify t. Here,

$$t' = t * scale\ of\ rating \quad (8)$$

Usually, the scale of rating is 5. Now, the new average rating is t = 0.714*5 = 3.57. Then, the value of parameter Trust, T = ((3.57*0.724)/5)*100 = 51.69%. Let, considering a customer has come to buy some products form Company A website. After completing all his prerequisites for completing a full transaction with this company, he needs to give his recommendation for different steps of 4 modules. Again, considering the previous states of Certain Trust Model, the present condition of Trust Values of the company are given in Table II.

TABLE II. Trust Values of Different Modules

| Module No | Variables | Values | Trust Values | Final Trust |
|---|---|---|---|---|
| Module 1 Existences | Physical Existence | c=0.6, t=3.5 | 42% | 43% |
| | People Existence | c=0.3, t=4 | 24% | |
| | Mandatory Existence | c=0.7, t=4.5 | 63% | |
| Module 2 Affiliation | Third Party Endorsement | C=0.6, t=4.25 | 51% | 70% |
| | Membership | C=0.9, t=4.8 | 86.4% | |
| | Portal | C=0.8, t=4.5 | 72% | |
| Module 3 Fulfillment | Delivery | c=0.5, t=4.35 | 43.5% | 59.5% |
| | Payment | c=0.8, t=4.30 | 69% | |
| | Community Customer | c=0.7, t=4.7 | 66% | |
| Module 4 Policy | Customer Satisfaction Policy | C=0.8, t=4.6 | 73.6% | 61% |
| | Privacy Statement | C=0.7, t=4.5 | 63% | |
| | Warranty Policy | C=0.5, t=4.6 | 46% | |

Now, the Trust Values for Merchant Trust for this transaction is T= ((43+70+59.5+61)/4) = 58.375%. So, considering other recommendation systems' trust values and previous trust values, the Trust of the company will be clustered in medium trust values class.

*B. Case Studies 2:*

Let us again consider a newer company, name B. This company uses the proposed model and tries to calculate their trust values for representing their Trustworthy condition to clients of the whole world. Considering all values and equations, they get the values shown in Table III.

TABLE III. Trust Values of Different Modules

| Module No | Variables | Values | Trust Values | Final Trust |
|---|---|---|---|---|
| Module 1 Existences | Physical Existence | c=0.5, t=3.8 | 38% | 57% |
| | People Existence | c=0.76, t=4.25 | 64.6% | |
| | Mandatory Existence | c=0.72, t=4.75 | 68.4% | |
| Module 2 Affiliation | Third Party Endorsement | C=0.5, t=4.5 | 45% | 39% |
| | Membership | C=0.9, t=4.8 | 86.4% | |
| | Portal | C=0.75, t=4.25 | 63.75% | |
| Module 3 Fulfillment | Delivery | c=0.46, t=4.45 | 41% | 50.5% |
| | Payment | c=0.6, t=4.7 | 56.4% | |
| | Community Customer | c=0.56, t=4.86 | 54% | |
| Module 4 Policy | Customer Satisfaction Policy | C=0.42, t=4.8 | 40% | 42.55% |
| | Privacy Statement | C=0.65, t=3.95 | 51.65% | |
| | Warranty Policy | C=0.55, t=3.26 | 36% | |

Now, the Trust Values for Merchant Trust for this transaction is T= ((57+39+50.5+42.55)/4) = 47.26%.

Now, Behavioral probability of company A is P = $(\frac{.58-.5}{.5})*100\%$ = 16% Up than base. Considering Base f=1.5. And Behavioral Probability of company B is P= $(\frac{.47-.5}{.5})*100\%$ = 6% lower than Base. Now, the newer client can easily distinguish between these two companies and can take decision easily with which company he will start his deal.

The advantages of proposed model over S.Nefti are given below:

TABLE IV.  Advantages of Proposed Model

| Points of Discussion | S.Nefti Model | Proposed Model |
|---|---|---|
| **Number of Fuzzy Classes** | 2-3 Membership function, So, lower specification | 5 Membership functions, So, classification and specification is Higher |
| **Dependency** | Depends on Uncertain Fuzzy System | Depends on Certain Fuzzy System |
| **Behavioral Probability** | Not Present | Present, so present condition can easily understandable |
| **Mapping in Cloud Architecture** | Not applicable | Fully Applicable |

## V. CONCLUSION

In this paper, we presented a system based on fuzzy logic based certain trust model to support the evaluation and the quantification of trust in E-commerce. As stated in many trust models, there are other aspects that contribute to the completion of online transactions. This includes the price, the rarity of the item and the experience of the customer. In this paper, we mainly concentrate on the structure of the E-commerce companies and different elements of it. With the help of proposed model, one can easily distinguish between companies and other elements of trust model will also be considered by the clients rating for the companies and for different transactions.

We have some new idea to imply in our proposed model in future. Firstly, we want to apply evolutionary algorithm with this model to optimize and design the rules. We want to apply price comparison as a parameter for a product in our model for ensuring accurate trust measuring model for a normal e-commerce website. Secondly, more development of behavioral probability parameter so that it can directly prohibit different types of security breaking questions like Sybil attack, false rating, etc.. At present, this parameter works indirectly with security options. Last of all, we want to work with all the points of E-Commerce architecture in our future model so that only our proposed model can fulfill all the requirements of customers who rely on online transactions for their business.

ACKNOWLEDGMENT

The authors wish to thank Sebastian Ries, Sheikh Mahbub Habib, Max Mühlhäuser and Vijay Varadharajan for their renownable work and thinking. Again, want to thank all the researchers and developers of Trust models for their excellent work which make authors work easier to accomplish.

REFERENCES

[1] Ries, S.: Trust in Ubiquitous Computing. PhD thesis, Technische University at Darmstadt,pp: 1-192, 2009
[2] K.S. Han and M.H. Noh, "Critical Failure Factors that Discourage the Growth of Electronic Commerce", International Journal of Electronic Commerce. 4(2), pp. 25-43, 1999.
[3] "Building an E-Commerce Trust Infrastructure SSL Server Certificates and Online Payment Services", Verisign, pg: 1-37
[4] Mohammed Alhamad, Tharam Dillon, and Elizabeth Chang A Trust-Evaluation Metric for Cloud applications *International Journal of Machine Learning and Computing, Vol. 1, No. 4, October 2011.*
[5] Jøsang, A., Ismail, R., Boyd, C.: A survey of trust and reputation systems for online service Provision. Decision Support Systems 43(2) (2007) 618–644.
[6] Vojislav Kecman, "Learning and Soft computing", Google online library, 2001.
[7] Kerr, R., Cohen, R.: Smart cheaters do prosper: defeating trust and reputation systems. In: AAMAS '09: Proceedings of The 8th International Conference on Autonomous Agents and Multiagent Systems, (2009) 993–1000.
[8] L. Zadeh. "Fuzzy sets", Journal of Information and Control, 8:338—353, 1965.
[9] D.W. Manchala, "E-Commerce Trust Metrics and Models", IEEE Internet Computing, March-April 2000, pp.36-44
[10] A. Jøsang, "Modelling Trust in Information Society", Ph.D. Thesis, Department of Telematics, Norwegian University of Science and Technology, Trondheim, Norway, 1998.
[11] Kawser Wazed Nafi, Tonny Shekha Kar, Md. Amjad Hossain, M. M. A. Hashem, "An Advanced Certain Trust Model Using Fuzzy Logic and Probabilistic Logic theory", IJACSA, vol 3, no 11, 2012.
[12] Samia, Nefti, FaridMeziane, KhairudinKasiran, A Fuzzy Trust Model for E-Commerce, 2004